\documentclass{INTERSPEECH2023}

\usepackage{makecell}
\usepackage[T1]{fontenc}
\usepackage[utf8]{inputenc}

\usepackage{mdframed}

\usepackage{latexsym}
\usepackage[T1]{fontenc}
\usepackage[utf8]{inputenc}
\usepackage{microtype}
\usepackage{inconsolata}

\usepackage{graphicx}

\usepackage{rotating}
\usepackage{multirow}
\usepackage{booktabs}
\usepackage{xspace}

\usepackage[subtle]{savetrees} 

\interspeechfinaltrue

\title{
From `Snippet-lects' to Doculects and Dialects: Leveraging Neural Representations of Speech for Placing Audio Signals in a Language Landscape}

\author{}

\name{Séverine Guillaume$^1$, Guillaume Wisniewski$^2$, Alexis Michaud$^1$}

\address{
  $^1$ Langues et Civilisations à Tradition Orale (LACITO), 
  CNRS~-- Université Sorbonne Nouvelle --
  Institut National des Langues et Civilisations Orientales (INALCO)  \\
  $^2$ Université de Paris Cité, Laboratoire de Linguistique Formelle (LLF), CNRS, F-75013 Paris, France}

\email{severine.guillaume@cnrs.fr,
  guillaume.wisniewski@u-paris.fr, alexis.michaud@cnrs.fr
}

\newcommand{\model}{\texttt{XLSR-53}\xspace}

\begin{document}
\maketitle
\begin{abstract}
\model, a multilingual model of speech, builds a vector representation from audio, which allows for a range of computational treatments. The experiments reported here use this neural representation 
    to estimate the degree of closeness between audio files, ultimately aiming to extract relevant linguistic properties. We use max-pooling to aggregate the neural representations from a `snippet-lect' (the speech in a 5-second audio snippet) to a `doculect' (the speech in a given resource), then to dialects and languages. 
    We use data from corpora of 11 dialects belonging to 5 less-studied languages. Similarity measurements between the 11 corpora bring out greatest closeness between those that are known to be dialects of the same language. The findings suggest that (i)~dialect/language can emerge among the various parameters characterizing audio files and (ii)~estimates of overall phonetic/phonological closeness can be obtained for a little-resourced or fully unknown language. 
    The findings help shed light on the type of information captured by neural representations of speech and how it can be extracted from these representations. 
\end{abstract}

\noindent\textbf{Index Terms}: pre-trained acoustic models, language documentation, under-resourced languages, similarity estimation

\section{Introduction}

The present research aims to contribute to a recent strand of research: exploring how pre-trained multilingual speech representation models like \model \cite{conneau21unsupervised} or \texttt{HuBERT} \cite{hubert} can be used to assist in the linguistic analysis of a language \cite{bartelds2022neural}. \model, a multilingual model of speech, builds a vector representation from an audio signal. The neural representation is different in structure from that of the audio recording. Whereas wav (PCM) audio consists in a vector of values in the range [-1:+1], at a bit-depth from 8 to 32 and a sampling rate on the order of 16,000~Hz, 
the \model neural representation contains 1,024 components, at a rate of 47 frames per second. The size of the vector representation is on the same order of magnitude as that of the 
audio snippet, and the amount of information can be hypothesized to be roughly comparable. But the neural representation, unlike the audio format, comes in a vector form that is tractable to a range of automatic treatments building on the vast body of work in data mining and machine learning.
The neural representation of speech holds potential for an epistemological turning-point comparable to the introduction of the spectrogram 8 decades ago \cite{potter_visible_1947,fulop2022,fant1960}. 

The experiments reported here use the neural representation yielded by \model (used off-the-shelf, without fine-tuning, unlike \cite{san21leveraging,guillaume-etal-2022-f}) as a means to characterize audio: estimating the degree of closeness between audio signals, and (ultimately) extracting relevant linguistic properties, 
teasing them apart from other types of information, e.g.\ technical characteristics of the recordings. We start out from 5-second audio snippets, and we pool neural representations (carrying out \textit{mean pooling}, i.e.\ averaging across frames) to progress towards the level of the entire audio file, then the entire corpus (containing several audio files). We thereby gradually broaden the scope of the neural representation from a `snippet-lect' (the speech present in an audio snippet\footnote{`Snippet-lect' is coined on the analogy of `doculect' \cite{good2013languoid}, to refer to the characteristics of a 5-second audio snippet.}) to a `doculect' (a linguistic variety as it is documented in a given resource \cite{good2013languoid}), then towards `dialects' (other groupings could also be used: by sociolect, by speaking style/genre, etc.) and, beyond, entire languages. 

In a set of exploratory experiments, we build neural representations of corpora of 11 dialects that belong to 5 under-resourced languages. We then 
use linguistic probes \cite{alain17understanding} (i.e.\ a multiclass classifier taking as input the frozen neural representation of an utterance and assigning it to a language, similarly to a language identification system) to assess the capacity of \model to capture language information. Building on these first results, we propose to use our probe on 
languages not present in the training set and to use its decisions as a measure of similarity between two languages, following the intuition that if an audio segment of an unknown language is identified as being of language \texttt{A}, then the language in the audio segment is “close” to \texttt{A}.

Representations like \model have already been used to develop language identification systems (e.g.\ \cite{tjandra22improved,fan21exploring}), but their use in the context of under-resourced languages and linguistic fieldwork datasets raises many challenges. First, there is much less data available for training and testing these systems both in terms of number of hours of audio and number of speakers. For instance, \texttt{VoxLingua}~\cite{valk21voxlingua}, a dataset collected to train language identification models, contains $6,628$ hours of recordings in 107 languages, i.e.\ at least an order of magnitude more data per language than typical linguistic fieldwork corpora. Second, the languages considered in a language documentation context have not been used for (pre-)training speech representations and have linguistic characteristics that are potentially very different from the languages used for pre-training them (on consequences of narrow typological scope for Natural Language Processing research, see \cite{bender2009}). The ability of models such as \model to correctly represent these languages remains an open question. We also aim to assess to what extent pre-trained models of speech can address these two challenges.

Similarity measurements between the 11 corpora bring out greatest closeness between those that are known to be dialects of the same language. Our findings suggest that dialect/language can emerge among the many parameters characterizing audio files as captured in \model representations (which also include acoustic properties of the environment, technical characteristics of the recording equipment, speaker ID, speaker gender, age, social group, as well as style of speech: speaking rate, etc.), and that there is potential for arriving at useful estimates of phonetic/phonological closeness. 
The encouraging conclusion is that, even in the case of a little-resourced or fully unknown language, `snippet-lects' and `doculects' can be placed relative to other speech varieties in terms of their closeness. 


An estimation of closeness between speech signals can have various applications. For computational language documentation \cite{michaud2018integrating,van2019future,prud2021automatic,guillaume-etal-2022-f}, there could be benefit in a tool for finding closest neighbours for a newly documented language (with a view to fine-tuning extant models for the newly documented variety, for instance), bypassing the need for explicit phoneme inventories, unlike in \cite{cotterell2017probabilistic}.
For dialectology, a discipline that traditionally relies on spatial models based on isogloss lines \cite{chagnaud2017shinydialect}, neural representations of audio signals for cognate words allow for calculating a phonetic-phonological distance along a dialect continuum \cite{bartelds2022neural}; our work explores whether cross-dialect comparison of audio snippets containing \textit{different utterances} also allows for significant generalizations. 
Last but not least, for the community of speech researchers, the task 
helps shed light on the type of information captured by neural representations of speech and how it can be extracted from these representations. This work is intended as a stepping-stone towards the mid-term goal of leveraging neural representations of speech to extract typological features from neural representations of speech signals: probing linguistic information in neural representations, to arrive at data-driven induction of typological knowledge \cite{ponti2019modeling}. Note that our work is speech-based, like \cite{suni2019comparative,de2022investigating}, and unlike text-based research predicting typological features (e.g.\ \cite{bjerva2017tracking}).

This article is organized as follows. In Section~\ref{sec:method} we introduce our system. 
In Section~\ref{sec:corpus} we briefly review the languages used in our experiments. Finally, we report our main experimental findings in Section~\ref{sec:res}.

\section{Probing Language Information in Neural Representations \label{sec:method}}

Predicting the language of a spoken utterance can, formally, be seen as a multi-class classification task that aims at mapping an audio snippet represented by a feature vector to one of the language labels present in the training set. Our implementation of this principle is very simple: we use 5-second audio snippets and use, as feature vector, the representation of the audio signal built by \model, a cross-lingual speech representation that results from pre-training a single Transformer model from the raw waveform of speech in multiple languages~\cite{conneau21unsupervised}. \model is a sequence-to-sequence model that transforms an audio file (a sequence of real numbers along the time dimension) into a sequence of vectors of dimension 1,024 sampled at 47~Hz  (i.e.\ it outputs 47~vectors for each second of audio). We use max-pooling to aggregate these vectors and map each audio snippet to a single vector. In all our experiments, we use a logistic regression (as implemented in the \texttt{sklearn} library \cite{scikit-learn}) as the multi-class classifier with $\ell_2$ regularization.

Importantly, our language identification system uses the representations built by \model without ever modifying them and is therefore akin to a linguistic probe \cite{alain17understanding}. We do not carry out fine-tuning of a pre-trained model. Language identification is a well-established task in the speech community and has been the focus of much research; our work does not aim at developing a state-of-the-art language identification model, but at showing that neural representations encode language information, and that this information can be useful for language documentation and analysis. Said differently, we do not aim to leverage ``emergent abilities'' of large language models \cite{schaeffer2023emergent}, but to explore one of their \textit{latent} abilities.

Our experimental framework allows us to consider several questions of interest to linguists. We can use various sets of labels, e.g.\ language names, or any level of phylogenetic (diachronic) grouping, or again typological (synchronic) groupings. We can also vary the examples the classifier is trained on. Among the many possibilities, we consider three settings:
\begin{itemize}
    \item a \emph{dialect identification} setting in which the classifier is trained on recordings of $N$ language varieties (dialects) and is then used (and evaluated) to recognize one of these;
    \item a \emph{language identification} setting which differs from the previous setting only by the definition of the label to predict: the goal is now to identify languages, which constitute groups of dialects. 
    Importantly, this classifier can be used to predict the language affiliation of a dialect that is not present in the train set, so that it can be used to predict, for instance, the language to which a hitherto unknown dialect belongs;
    \item a \emph{similarity identification} setting which differs from the first setting only by the definition of the train set: in this setting, we use our model on utterances of a dialect that is not present in the train set. Since the classifier cannot predict the exact dialect (as its label is not available from within the train set), it seems intuitively likely to choose the label of a dialect with similar characteristics. Crucially, we believe that this setting will therefore allow to identify similarities between language varieties.
\end{itemize}

\section{Information on Languages and Dialects\label{sec:corpus}}

In all our experiments, we use datasets from the Pangloss Collection~\cite{michailovsky14documenting},\footnote{Website: \href{https://pangloss.cnrs.fr/?mode=pro&lang=en}{pangloss.cnrs.fr}. A tool for bulk downloads and for tailoring reference corpora is available: \href{https://gitlab.com/lacito/outilspangloss}{OutilsPangloss}.} 
an open archive of 
(mostly) endangered languages.
Our experiments focus on 11 dialects that belong to five languages:
\begin{itemize}
    \item two dialects of Nepali: 
    Achhami (Glottocode \cite{hammarstrom2015glottolog}: doty1234) 
    and Dotyal (doty1234);
    \item two dialects of Lyngam (lyng1241): Langkma and Nongtrei;
    \item three varieties of Na-našu, a dialect of Shtokavian Serbo-Croatian (shto1241) spoken by Italian Croats;
    \item two dialects of War (khas1268): Amwi (warj1242) and Nongtalang (nong1246);
    \item two dialects of Na (yong1270): Lataddi Na (lata1234) and Yongning Na (yong1288).
\end{itemize}
We also consider two additional languages, Naxi (naxi1245) and Laze (laze1238), 
because of their closeness to Na \cite{jacques11approaching}. 


For the sake of consistency in the experiments reported here, we use “dialect” as the lowest-level label, and “language” for the first higher level, as a convention.
We are aware that the distance between “dialects” (and between “languages”) varies significantly from one case to another. We do not assume that the distance between Achhami and Dotyal (dialects of Nepali) is (even approximately) the same as that between Langkma and Nongtrei (dialects of Lyngam), or between Lataddi Na and Yongning Na. The key assumption behind our use of terms is that language varieties referred to as ``dialects'' of the same language are close enough that it makes sense to assume that the degree of phonetic similarity between them can serve as a rule-of-thumb estimate for the distance that separates them, without requiring higher-level linguistic information (of the type used to train a language model).

In this preliminary study we have decided to focus on a small number of languages and to focus on qualitative analysis of our results, rather than running a large-scale experiment on dozens of languages. The languages are chosen according to the size of the available corpora and specific properties. We favored continuous speech (we left aside corpora consisting solely of word lists or materials elicited sentence by sentence). 

For each of these languages we extracted 2 to 50 files of variable length (from 33~seconds to 30~minutes). 

\section{Experiments \label{sec:res}}

In all our experiments, we evaluate the capacity of our classifier to predict the correct language information (either the label of a specific dialect or the name of a language) using the usual metrics for multi-class classification, namely, precision, recall and their combination in the $F_1$ score.

\textbf{Dialect Identification} To test the ability of a classifier to recognize a dialect from the representations built by \model, we consider a classifier using the names of the 13 dialects or languages described in Section~\ref{sec:corpus} as its label set.  We try out two configurations. In the first one, all the utterances of a dialect are randomly divided into a test set (20\% of utterances) and a training set (80\%). In the second configuration, the training corpus is made up of 80\% of the files of a dialect and the test corpus contains the remaining 20\%. While the latter configuration is closer to the real conditions of use of our system~(guaranteeing that the utterances of the test corpus come only from files that have not been seen at training), it is more difficult to control the size of the train and test sets, which makes the analysis less straightforward. 

The results are reported in Table~\ref{tab:res_id}. They show that, in both configurations, a simple classifier is able to identify the correct dialect label for an utterance with high accuracy, showing that \model representations encode language information. Similar observations have already been reported (see, e.g., \cite{deseyssel22probing}), but to the best of our knowledge, our work is the first evaluation of the capacity of \model representations to identify under-documented language varieties whose characteristics are potentially very different from the languages seen at (pre-)training~\cite{guillaume-etal-2022-f}. Interestingly, the quality of predictions does not seem to be influenced by the amount of training data (a similar paradox is reported in the evaluation of another large language model in multilingual learning: ChatGPT \cite{lai2023chatgpt}).

\begin{table}[!tb]
    \centering
    \resizebox{8.3cm}{!}{
    \begin{tabular}{l c lll c lll}
    \toprule
      && \multicolumn{3}{c}{utterance split} && \multicolumn{3}{c}{file split}  \\
      \cline{3-5} \cline{7-9}
      && precision & recall & $F_1$          && precision & recall & $F_1$ \\ \midrule
    Achhami               && 0.98   &   0.91   &   0.95 && 0.88   &   0.93   &   0.90 \\
    Dotyal                && 1.00   &   0.98   &   0.99 && 1.00   &   0.30   &   0.46 \\
    Laze                  && 0.96   &   0.98   &   0.97 && 0.80   &   0.96   &   0.87 \\
    Langkma               && 0.89   &   0.90   &   0.90 && 0.74   &   0.96   &   0.83 \\
    Nongtrei              && 1.00   &   1.00   &   1.00 && 1.00   &   1.00   &   1.00 \\
    Acquaviva Collecroce  && 0.93   &   0.88   &   0.90 && 0.80   &   0.95   &   0.87 \\
    Montemitro            && 0.92   &   0.91   &   0.92 && 0.94   &   0.80   &   0.87 \\
    San Felice del Molise && 0.89   &   0.97   &   0.93 && 0.87   &   0.87   &   0.87 \\
    Naxi                  && 0.99   &   0.96   &   0.97 && 0.85   &   0.95   &   0.90 \\
    Lataddi Na             && 0.97   &   0.98   &   0.97 && 0.93   &   0.96   &   0.94 \\
    Amwi                  && 0.94   &   0.93   &   0.93 && 0.67   &   0.89   &   0.76 \\
    Nongtalang            && 0.93   &   0.95   &   0.94 && 0.90   &   0.84   &   0.87 \\
    Yongning Na           && 0.97   &   0.97   &   0.97 && 0.95   &   0.97   &   0.96 \\
    \midrule
    \textbf{macro average}  && 0.95 &   0.95   &   0.95 && 0.87   &   0.88   &   0.86 \\
    \bottomrule
    \end{tabular}}
    \caption{Result of our dialect identification experiments. “Utterance split” refers to the setting in which data from the same file can appear both in the train and test sets. “File split” corresponds to the setting in which we require that the files of the train and test sets are different.\label{tab:res_id}}
    \centering
\end{table}

The recordings considered in the experiment we have just described were all collected in the context of linguistic fieldwork, and thus have some peculiarities that may distort the conclusions we have just drawn. In particular, most of the dialects we considered have recordings of a single speaker. Moreover, different dialects of the same language were often recorded by the same linguist, using the same recording setup (in particular, the same microphone). We therefore need to check whether our classifiers just learn to distinguish speakers (in many cases: one per dialect) or recording conditions. In order to rule out this possibility, we carried out a control experiment in which we tried to predict the file name (serving as proxy information for the speaker and the recording conditions). A logistic regression trained in the 80-20 condition described above achieved a macro $F_1$ score of 0.45, showing that the decision of the classifier is largely based on linguistic information, not solely on information about the recording conditions. 

\begin{table}[t!!]
    \begin{tabular}{lccc}
        \toprule
        & precision & recall & $F_1$ \\
        \midrule
        Laze$^\dag$    &   0.97   &   0.98  &    0.98 \\
      Lyngam    &   1.00   &   0.99  &    0.99 \\
          Na    &   0.96   &   0.99  &    0.97 \\
     Na-našu    &   0.99   &   0.99  &    0.99 \\
        Naxi$^\dag$    &   0.89   &   0.98  &    0.93 \\
      Nepali    &   1.00   &   0.46  &    0.63 \\
         War    &   0.89   &   0.96  &    0.93 \\
         \midrule
         \textbf{macro average} & 0.96   &   0.91   &   0.92 \\
        \bottomrule
    \end{tabular}
    \centering
    \caption{Performance of a classifier trained to predict languages (group of dialects). Languages consisting of a single dialect are indicated with a $\dag$. \label{tab:res_group}}

    \centering
    \begin{tabular}{l c ccc}
    \toprule
                && precision & recall & $F_1$ \\
    \midrule
      Lyngam    &&   0.59  &    0.81   &   0.68 \\ 
          Na    &&   0.86  &    0.83   &   0.84 \\ 
     Na-našu    &&   0.48  &    0.75   &   0.59 \\ 
      Nepali    &&   0.09  &    0.09   &   0.09 \\ 
         War    &&   0.74  &    0.60   &   0.66 \\ 
    \midrule
   \textbf{macro average}  &&  0.55  &    0.62  &    0.57 \\
   \bottomrule
\end{tabular}
\caption{Performance of a classifier trained to predict the language (group of dialects) of dialects not seen during training. Naxi and Laze have been left out as there is a single variety of these languages in our dataset.\label{tab:leave_one_out}}
\end{table}

\begin{table*}[t!!]
    \centering
    {\footnotesize
    \begin{tabular}{llrrrrrrrrrrrrr}
        \toprule
         &  & \rotatebox{90}{Laze} & \rotatebox{90}{Langkma} & \rotatebox{90}{Nongtrei} & \rotatebox{90}{Lataddi Na} & \rotatebox{90}{Yongning Na} & \rotatebox{90}{Acquaviva C.} & \rotatebox{90}{Montemitro} & \rotatebox{90}{San Felice dM} & \rotatebox{90}{Naxi} & \rotatebox{90}{Achhami} & \rotatebox{90}{Dotyal} & \rotatebox{90}{Amwi} & \rotatebox{90}{Nongtalang} \\
        \midrule
        \multicolumn{3}{l}{\textit{Laze}} \\
         & Laze & --- & 0.3 & 0.0 & 1.7 & \bfseries 35.0 & 9.2 & 6.6 & 0.0 & 17.9 & 0.1 & 0.1 & 2.4 & 26.7 \\
        \multicolumn{3}{l}{\textit{Lyngam}} \\
         & Langkma & 6.2 & --- & 0.0 & 0.6 & 11.9 & 0.0 & 0.0 & 3.7 & 4.2 & 2.5 & 0.0 & 5.4 & \bfseries 65.5 \\
         & Nongtrei & 0.0 & 0.0 & --- & 0.0 & 0.0 & 0.0 & 0.0 & 0.0 & 0.7 & 0.0 & 0.0 & 0.0 & \bfseries 99.3 \\
         \multicolumn{3}{l}{\textit{Na}} \\
         & Lataddi Na & 0.6 & 0.4 & 0.0 & --- & \bfseries 52.1 & 1.3 & 0.5 & 0.8 & 28.5 & 1.6 & 0.2 & 3.5 & 10.6 \\
         & Yongning Na & 4.4 & 0.8 & 0.0 & \bfseries 72.4 & --- & 0.0 & 1.5 & 0.0 & 12.5 & 0.3 & 0.0 & 2.1 & 5.8 \\
         \multicolumn{3}{l}{\textit{Na-našu}} \\
         & Acquaviva Collecroce & 0.5 & 0.5 & 0.0 & 4.0 & 0.2 & --- & \bfseries 64.2 & 4.2 & 8.6 & 0.0 & 0.0 & 3.0 & 14.8 \\
         & Montemitro & 0.7 & 0.0 & 0.0 & 0.0 & 0.7 & \bfseries 84.4 & --- & 6.4 & 4.7 & 0.0 & 0.0 & 0.0 & 3.1 \\
         & San Felice del Molise & 3.3 & \bfseries 28.2 & 0.0 & 1.5 & 0.9 & 11.0 & 7.7 & --- & 19.3 & 23.4 & 0.0 & 1.5 & 3.3 \\
         \multicolumn{3}{l}{\textit{Naxi}} \\
        & Naxi & 16.5 & 1.4 & 0.2 & 13.0 & 8.3 & 0.3 & 5.7 & 5.0 & --- & 0.6 & 1.3 & 4.0 & \bfseries 43.7 \\
        \multicolumn{3}{l}{\textit{Nepali}} \\
        & Achhami & 0.6 & 0.3 & 0.0 & 3.4 & 0.3 & 0.0 & 0.0 & 16.9 & 20.2 & --- & 0.3 & 8.3 & \bfseries 49.7 \\
         & Dotyal & 0.0 & 0.0 & 0.0 & 20.0 & 0.0 & 0.0 & 0.0 & 0.0 & \bfseries 35.0 & 0.0 & --- & \bfseries 35.0 & 10.0 \\
        \multicolumn{3}{l}{\textit{War}} \\
         & Amwi & 0.0 & 0.1 & 0.0 & 9.6 & 17.0 & 1.7 & 0.1 & 1.1 & 7.0 & 5.7 & 0.0 & --- & \bfseries 57.5 \\
         & Nongtalang & 5.7 & 3.2 & 2.8 & 8.6 & 8.4 & 5.4 & 1.4 & 5.0 & 21.2 & 5.8 & 0.1 & \bfseries 32.5 & --- \\
        \bottomrule
        \end{tabular}
    }
    \centering
    \caption{Distribution of the labels predicted by a classifier trained on 12 dialects (in columns) and used on a 13\textsuperscript{th} dialect (unseen at training). 
    Thus a classifier trained on all except Yongning Na identifies 72.4\% of Yongning Na utterances as Lataddi Na and 12.5\% as Naxi. \label{tab:res_similarity}}
\end{table*}

\textbf{Language Identification} In a second experiment, we test the ability of our classifier to identify languages (that is, groups of dialects). We consider, again, two conditions to train and evaluate our classifier. In a first condition, the train and test sets are randomly sampled from all the recordings we consider (with the usual 80\%-20\% split) without any condition being imposed on the files or languages. All dialects are therefore present in both the test and train sets. In a second condition, the test set is put together by selecting, for each language (group of dialects), all the recordings of a randomly chosen dialect. The test set is thus made up of 5 dialects that have not been seen at training.

Table~\ref{tab:res_group} reports results in the first condition. The classifier succeeds in identifying the correct language in the vast majority of cases, a very logical result since the same languages are present in the train and test sets and the experiments reported in the previous paragraph proved that it is possible to identify dialects with good accuracy. To verify that the classifier was able to extract linguistic information rather than merely memorizing arbitrary associations between dialects, we performed a control experiment in which we divided the 13 dialects into 5 arbitrary groups having the same size as the languages (dialect groups) considered in the previous experiment. A classifier considering these groups as labels achieves a macro $F_1$ score of 0.85. While this score is high, it is notably lower than the score obtained by predicting linguistic families, showing that the classifier decisions are, to a significant extent, based on linguistic criteria.


Table~\ref{tab:leave_one_out} shows the results for the second condition, in which we evaluate the capacity of a classifier to predict the language (dialect group) of a dialect that was not part of the train set. Scores vary greatly by language (group of dialects) and several factors make it difficult to interpret these results. First, removing a dialect completely from the train set can result in large variation in its size and the results of Table~\ref{tab:leave_one_out} are not necessarily comparable with those reported so far. Second, some confounders seem to cause particularly poor performance for certain groups of dialects. For example, recordings of Dotyal are mainly sung epic poetry, so it is not surprising that any generalization across the two dialects of Nepali is difficult. Gender seems to be another confounder: several corpora only contain recordings by speakers of the same gender, and a quick qualitative study seems to show that a model trained on a female speaker does not perform well on data by a male speaker (and conversely). Note, however, that our evaluation of the performance of the classifier puts it at a disadvantage since it is evaluated at the level of a 5-second snippet and not of an entire recording. It is not unlikely that the performance would be better if we predicted a single label for a whole recording (for example by taking the most frequent label among those of all snippets).

\textbf{Similarity Identification Setting} In our last experiment, we trained 12 classifiers, considering all dialects but one for training and looking at the distribution of predicted labels when the classifier had to identify snippets of the held-out language. As explained in Section~\ref{sec:method}, the classifier cannot predict the correct label (since the target language is not present in the training corpus) but might, we believe, pitch on a language with similar characteristics. Results of this experiment are reported in Table~\ref{tab:res_similarity}. They allow us to draw several interesting conclusions. 

First, these results show that the classifier pitches consistently on one and the same label. In almost every case, the distribution of predicted labels is concentrated on a few labels. That means that the classifier typically identifies almost all snippets from an audio file as being in the same language. Second, in several cases (e.g.\ for dialects of Na, War or Na-našu), the classifier recognizes the unknown language as a dialect of the same group: for instance Yongning Na utterances are mainly labelled as Lataddi Na (the dialect of a nearby village). In addition to its interest for the automatic identification of dialect groups, this observation proves that \model uncovers representations that somehow generalize over small dialectal variation. 

Further experiments are needed to understand the two cases where the output of the classifier disagrees with the gold-standard clustering: the San Felice del Molise dialect of Na-našu, and the two dialects of Lyngam (Langkma and Nongtrei). (For Nepali, a plausible confounder was mentioned above: data type~--~genre~--, as the Dotyal corpus consists of sung epics.)

\section{Conclusions}

Our exploratory experiments exploring the capacity of \model to place audio signals in a language and dialect landscape confirm the interest of neural representations of speech as an exciting avenue of research. Further work is required to ensure that a dialect identification system bases its decisions on phenomena (detecting relevant phonetic-phonological structures), not on parameters such as recording conditions, speaker characteristics (gender, age...) and speech genre/style, which constitute confounders in a language identification task. 
In future work, we plan to reproduce the experiments on corpora of better-resourced languages, such as LibriVox or CommonVoice, for which it is easier to control recording conditions, speaker gender, and the amount of training data.


\bibliographystyle{IEEEtran}
\bibliography{snippet_lects}

\end{document}